# A distributed Approach for Access and Visibility Task under Ergonomic Constraints with a Manikin in a Virtual Reality Environment

F. Bidault(1), D. Chablat(1), P. Chedmail(1), L. Pino(2)
(1)Institut de Recherche en Communications et Cybernétique de Nantes∗,
1, rue de la Noë, B.P. 92101, 44321 Nantes Cedex 3 France
(2)Laboratoire d'Informatique Industrielle, Ecole Nationale d'Ingénieurs de Brest,
Parvis Blaise Pascal, 29280 Plouzané B.P. 30815 - 29608 Brest Cedex France
E-mail Patrick.Chedmail@irccyn.ec-nantes.fr

**Abstract**

*This paper presents a new method, based on a multi-agent system and on digital mock-up technology, to assess an efficient path planner for a manikin for access and visibility task under ergonomic constraints. In order to solve this problem, the human operator is integrated in the process optimization to contribute to a global perception of the environment. This operator cooperates, in real-time, with several automatic local elementary agents. The result of this work validates solutions brought by digital mock-up and that can be applied to simulate maintenance task.*

## 1 Introduction

In an industrial environment, the access to a sharable and global view of the enterprise project, product, and/or service appears to be a key factor of success. It improves the triptych delay-quality-cost but also the communication between the different partners and their implication in the project. For these reasons, the digital mock-up (DMU) and its functionality are investigated more deeply by industrials. Based on computer technology and virtual reality, the DMU consists in a platform of visualisation and simulation that can cover different processes and areas during the product lifecycle such as product conception, industrialisation, production, maintenance, recycling and/or customer support (fig. 1).

The digital model enables the earlier identification of possible issues and a better understanding of the processes even, and maybe above all, for actors who are not specialists. Thus, a digital model allows decisions to be made before expensive physical prototypes have been built. Even if evident progresses were noticed and applied in the domain of DMUs, significant progresses are still awaited for a placement in an industrial context. As a matter of fact, the digital model offers a way to explore areas such as maintenance or ergonomics of the product that were traditionally ignored at the beginning phases of a project; new processes must consequently be developed.

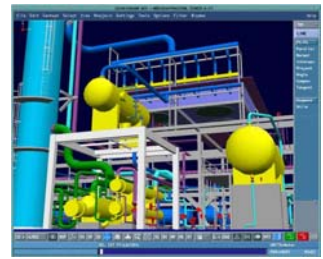

**Figure 1**. Manufacturing simulation.

Through the integration of a manikin in a virtual environment, the suitability of a product, its form and functions can be assessed. Moreover, when simulating a task that should be performed by an operator with a virtual manikin model, feasibility, access and visibility can be checked. The conditions of the performances in terms of efforts, constraints and comfort can also be analyzed. Modifications on the product or the task itself may follow but also a better and earlier formation of the operators to enhance their performances in the real environment. Moreover, such a use of the DMU leads to a better conformance to health and safety standards and to a maximization of human comfort and safety.

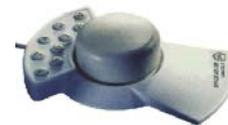

**Figure 2.** SpaceMouse (LogitechTM).

With virtual reality tools such as 3D manipulators (fig. 2), it is possible to manipulate directly an object in a cluttered environment. Some drawbacks are the difficulty to manipulate the object with as ease as in a real environment, due to the lack of kinematics constraints and the automatic collision avoidance. As a matter of fact, interference detection between parts is often displayed

---

through color changes of parts in collision but collision is not avoided.

Another approach consists in integrating automatic functionality into the virtual environment in order to ease the user's task. Many research topics in the framework of robotics dealing with the definition of collision-free trajectories for solid objects are also valid in the DMU. Some methodologies need a global perception of the environment, like (1) visibility graphs proposed by Lozano-Pérez and Wesley [1], (2) geodesic graphs proposed by Tournassoud [2], or (3) Voronoï's diagrams [3]. However, these techniques are very CPU consuming but lead to a solution if it exists. Some other methodologies consider the moves of the object only in its close or local environment. The success of these methods is not guaranteed due to the existence of local minima. A specific method was proposed by Khatib [4] and enhanced by Barraquand and Latombe [5]. In this method, Khatib's potentials method is coupled with an optimization method that minimizes the distance to the target and avoid collisions. All these techniques are limited, either by the computation cost, or the existence of local minima as explained by Namgung [6]. For these reasons a designer, is required in order to validate one of the different paths found or to avoid local minima.

The accessibility and the optimum placement of an operator to perform a task is also a matter of path planning that we propose to solve with DMU. In order to shorten time for a trajectory search, to avoid local minima and to suppress tiresome on-line manipulation, we intend to settle for a mixed approach of the above presented methodologies. Thus, we use local algorithm abilities and global view ability of a human operator, with the same approach as [7]. Among the local algorithms, we present these ones contributing to a better visibility of the task, in term of access but also in term of comfort.

## 2 Multi-agent systems

The above chapter points out the local abilities of several path planners. Furthermore, human global vision can lead to a coherent partition of the path planning issue. We intend to manage simultaneously these local and global abilities by building an interaction between human and algorithms in order to have an efficient path planner [8] for the manikin with respect of ergonomic constraints.

### 2.1 History

Several studies about co-operation between algorithm processes and human operators have shown the great potential of co-operation between agents. First concepts were proposed by Ferber [9]. These studies led to the creation of a "Concurrent Engineering" methodology based on network principles, interacting with cells or modules that represent skills, rules or workgroups. Such studies can be linked to work done by Arcand and Pelletier [10] for the construction of a cognition based multi-agent architecture. This work presents a multi-agent architecture with human and society behavior. It uses cognitive psychology results within a co-operative human and computer system.

All these studies show the important potential of multi-agent systems (MAS). Consequently, we built a manikin "positioner", based on MAS, that combines human interactive integration and algorithms.

### 2.2 Retained multi-agent theory

Several workgroups have established rules for the definition of the agents and their interactions, even for dynamic architectures according to the environment evolution [9, 11]. From these analyses, we keep the following points for an elementary agent definition. An elementary agent:
- is able to act in a common environment,
- is driven by a set of tendencies (goal, satisfaction function, etc.),
- has its own resources,
- can see locally its environment,
- has a partial representation of the environment,
- has some skills and offers some services,
- has behavior in order to satisfy its goal, taking into account its resources and abilities, according to its environment analysis and to the information it receives.

The points above show that direct communications between agents are not considered. In fact, our architecture implies that each agent acts on its set of variables from the environment according to its goal.

### 2.3 Correlation between path planning and MAS

The method used in automatic path planners is schematized fig. 3a. A human global vision can lead to a coherent partition of the main trajectory as suggested in [12]. Consequently, another method is the integration of an operator to manage the evolution of the variables, taking into account his or her global perception of the environment (fig. 3b). To enhance path planning, a coupled approach using multi-agent and distributed principles as it is defined in [8] can be build; this approach manages simultaneously the two, local and global, abilities as suggested fig. 3c. The virtual site enables graphic visualization of the database for the human operator, and communicates positions of the virtual objects to external processes.

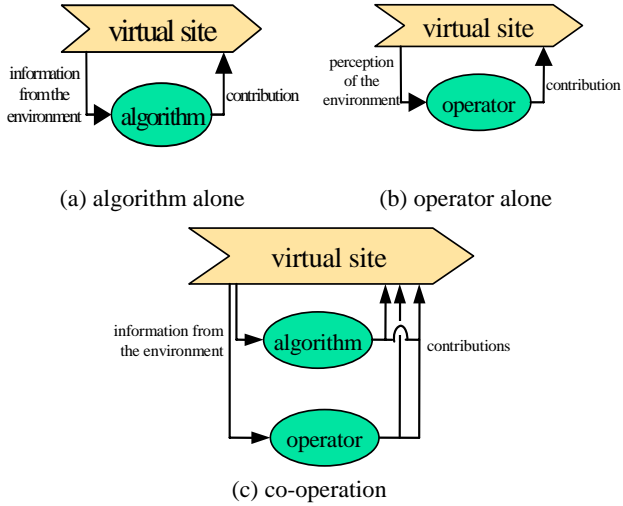

**Figure 3.** Co-operation principles.

As a matter of fact, this scheme is to correlate with the architecture of the so-called system "blackboard". This principle is described in [9, 13, 11]. A schematic presentation is presented on fig. 4. The only medium between agents is the common database of a virtual reality environment. The human operator can be consider as an elementary agent for the system, co-operating with some other elementary agents that are simple algorithms.

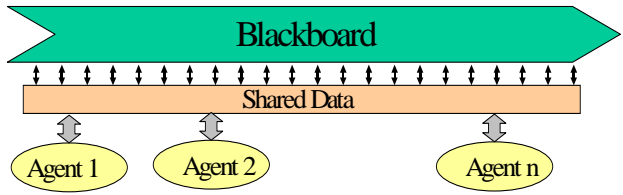

**Figure 4.** Blackboard principle with co-operating agents.

### 2.4 Considered approach

The approach we retained is the one proposed in [7] whose purpose was to validate new CAD/CAM solutions based on a distributed approach using a virtual reality environment. This method has successfully demonstrated its advantage by enabling to prove, with a reduced time, the montage of different elements for which ones it was before necessary to build real and physical mock-ups. We kept the same architecture and developed some elementary agents for the manikin (fig. 5). In fact, each agent can be also divided in elementary agents.

Each agent acts at a specific time sampling which is pre-defined by a specific rate of activity $\lambda_i$. When acting, the agent sends a contribution, normalized by a value $\Delta$, to the environment and/or the manipulated object (the manikin in our study). In fig. 6, we represent the agent *Collision* with a rate of activity equal to 1, the agent *Attraction* has a rate of 3 and agents *Operator* and *Manikin* a rate of 9. This periodicity of the agent actions is a characteristic of the architecture: it expresses a priority between each of the goals of the agents. To supervise each agent activity, we use an event programming method where the main process collects agent contributions and updates the database [7]. The normalization of the actions of the agents (the value $\Delta$) induces that the actions are relative and not absolute.

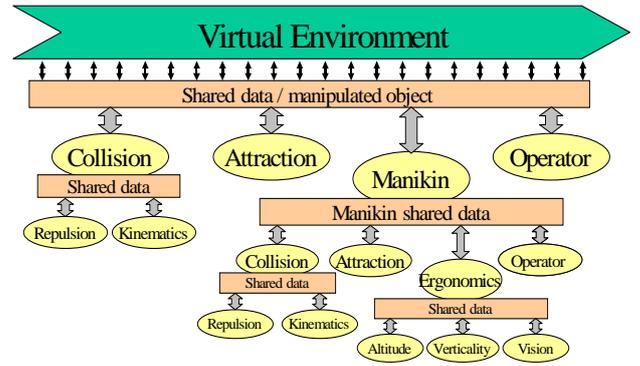

**Figure 5.** Co-operating agents in the retained approach.

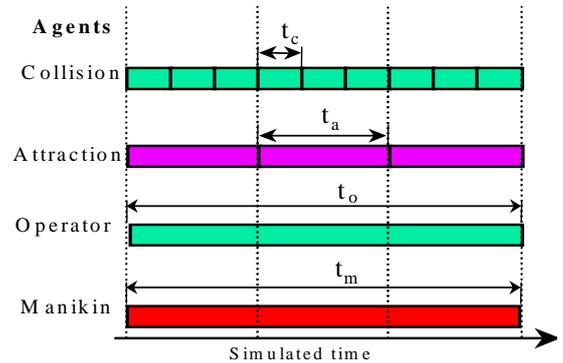

**Figure 6.** Time and contribution sampling.

## 3 Visibility and maintainability check with multi-agent system in virtual reality

### 3.1 Introduction

For the visibility check, we focus our work on the trunk and the head configurations. The joint coupling the head with the trunk is characterized by three rotations $\alpha_b$, $\beta_b$ and $\theta_b$ whose range limits are defined by ergonomic constraints (fig. 7). This data can be found using the results of ergonomic research [14]. To solve the problem of visibility, we define a cone C whose vertex is centered between the two eyes and whose base is located in the plane orthogonal to **u**, centered on the target (fig. 8). The cone width $\varepsilon_c$ is variable.

Thus, additionally to the position and orientation variables of all parts in the cluttered environment (including the manikin itself), we consider in particular:

➢ Three degrees of freedom for the manikin to move in the x-y plane: $\mathbf{x}_m = (x_m, y_m, \theta_m,)^t$. It is also possible to take into account a degree of freedom $z_m$ if we want to give to the manikin the capacity to clear an obstacle.

➢ Three degrees of freedom for the head articulation to manage the manikin vision: $\mathbf{q}_b = (\alpha_b, \beta_b, \theta_b)^t$ with their corresponding joint constraints.

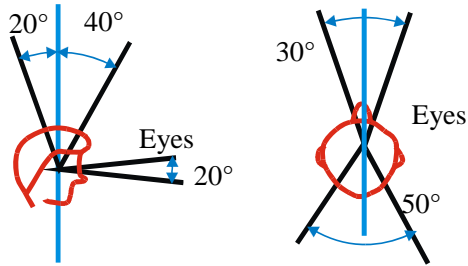

**Figure 7.** Example of joint limits and visibility capacity.

The normalized contributions from the agents are defined with two fixed parameters: $\Delta_{pos}$ for translational moves and $\Delta_{or}$ for rotational moves.

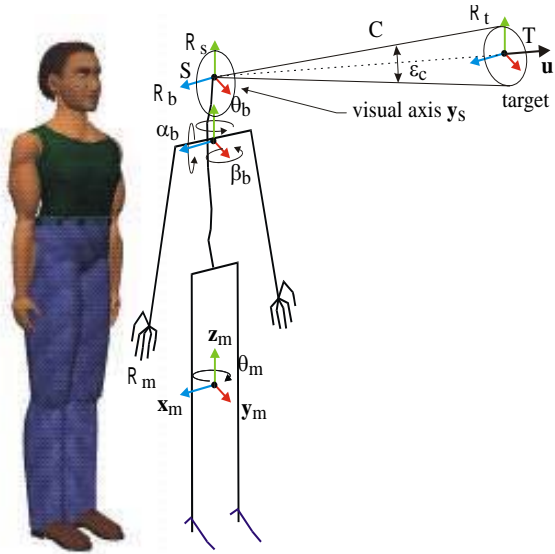

**Figure 8.** Manikin skeleton, visibility cone and target definition.

### 3.2 Agents ensuring task visibility and comfort

We present below all the elementary agents used in our system to solve the access and visibility task.

- *Attraction* agent for the manikin

The goal of the *attraction* agent is to enable the manikin to reach the target with the best trunk posture, that is:

➢ To orient the projection of $\mathbf{y}_m$ on the floor plane collinear to the projection of **u** on the same plane by rotation of $\theta_m$ (fig. 8),

➢ To position $x_m$ and $y_m$, coordinates of the manikin in the environment floor, as close as possible to the target position (fig. 8),.

This *attraction* agent only considers the target and does not take care of the environment. This agent is similar to the attraction force introduced by Khatib [4], and gives the required contributions $x_{att}$, $y_{att}$, and $\theta_{att}$ according to the attraction toward a target referenced as above. These contributions, which act on the manikin leading member position and orientation (in our case the trunk), are normalized according to $\Delta_{pos}$ and $\Delta_{or}$.

- *Repulsion* agent between manikin and the cluttered environment

This *repulsion* agent acts in order to avoid the collisions between the manikin and the cluttered environment, which is fixed for our study.

Several possibilities can be used in order to build a collision criterion. The intersection between two parts A and B in collision, as shown by fig. 9a, can be quantified in several ways. We can consider either volume V of collision, or the surface $\Sigma$ of collision or the depth $D_{max}$ of collision (fig. 9b). The main drawback of these approaches comes from the difficulty to determine these values. Moreover, 3D topological operations are not easy because our virtual reality software uses polyhedral surfaces to define 3D objects. To determine $D_{move}$, the distance to avoid the collision (fig. 9b), we have to store old positions of the parts, so this quantification does not only uses the database at a given instant but uses former information. This solution cannot be kept for our blackboard architecture that only provides global environment status at an instant.

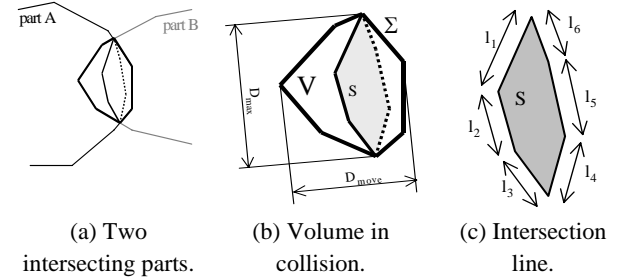

(a) Two intersecting parts.  (b) Volume in collision.  (c) Intersection line.

**Figure 9.** Collision criteria.

Another quantification of the collision is possible with the use of the collision line between the two parts. With this collision line, we can determine the maximum surface S or the maximum length of the collision line $l = \Sigma\ l_i$ (fig. 9c) and the gradient of collision length according to the Cartesian environment frame. The second method is the one we adopted in our system in reasons of the computational time performance. It is coupled with a method based on sphering box collision checking to enhance the algorithm. The line of collision is computed between the environment and all the manikin's members.

From the gradient vector of the collision length $\mathbf{grad}_{(x,y,\theta)}(l)$, contributions $x_{rep}$, $y_{rep}$, and $\theta_{rep}$ are computed by the *repulsion* agent. These contributions, acting on the manikin trunk position and orientation, are normalized according to $\Delta_{pos}$ and $\Delta_{or}$.

- Head orientation agent

The goal of the *head orientation* agent is to orient the head of the manikin to see the target ensuring the optimum configuration that maximizes visual comfort. Finding the optimum configuration consists into minimizing efforts on the joint coupling the head with the trunk and minimizing ocular efforts. We simplify the problem by considering that the manikin has a monocular vision, defined by a cone whose principal axis, called vision axis, is along $\mathbf{y}_s$ and whose vertex is the center of manikin eyes. If the target belongs to the vision axis, ocular efforts are considered null. Our purpose is then to orient $\mathbf{y}_s$ collinear to $\mathbf{u}$ by rotation of $\alpha_b$ and $\theta_b$ (fig. 8), with respect of joint limits, which are parameters that can be changed through the software. A joint limit average for an adult is given in fig. 7.

The algorithm of this agent is similar to the *attraction* agent algorithm presented above; contributions $\alpha_{head}$ and $\theta_{head}$, after normalization, are applied to the joint coupling the head to the manikin trunk.

- *Visibility* agent

The *visibility* agent ensures that the target is visible, that is, that no interference occurs between the segment ST, linking the center of manikin eyes and the target, and the cluttered environment. Rather than a segment, we consider the collisions between a facetted simplification of the cone C (fig. 8), and the environment. The repulsion algorithm is exactly the same as the one presented above:

➢ we determine the collision line length,
➢ if non equal to zero, normalized contributions are determined from $x_{vis}$, $y_{vis}$, and $\theta_{vis}$ computed by the *visibility* agent according to the gradient vector of the collision length,
➢ contributions are applied to the manikin trunk.

It is to notice that some contributions may also be applied to the head orientation since by turning the head, collisions between the simplified cone with the environment may also occur.

The use of a simplified cone offers the advantage of combining an ergonomic criterion with the repulsion effect. As a matter of fact, when the vision axis $\mathbf{y}_s$ is inside the cone C (fig. 8), we widen the cone, respecting a maximum limit. If not, we decrease its vertex angle, also with respect of a minimum limit that corresponds to the initial condition when starting this *visibility* agent. The maximum limit may be expressed according to the target size or/and to the type of task to perform: proximal or distant visual checking, global or specific area to control.

- *Operator* agent on the manikin

One of the aims of the study is to integrate a human operator within the MAS to operate in real-time. The *operator* has a global view of the cluttered environment displayed by means of the virtual reality software. Her or his action must be simple and efficient. For this, we use a device (SpaceMouse) that allows us to manipulate a body with six degrees of freedom.

The action of the *operator* agent only considers the move of the leading object, which is in our case the manikin trunk. Parameters come from position $x_{op}$ and $y_{op}$ and orientation $\theta_{op}$ returned from the device. These contributions are normalized, as for the *attraction* or *repulsion* agents.

## 4 Results and conclusions

This method is under testing to check the visual accessibility of specific elements under a trap of an aircraft. The digital model is presented in fig.10 and the list of elementary agents is depicted in the master agent window in fig. 11. In this example, the *repulsion* agent for the manikin (**Repulsion**), the *visibility* agent (**Visual**) and the *head orientation* agent (**Cone**) have a specific rate of activity equal to 1, meaning that their actions have priority but it is possible for the operator to change in real-time this activity rate. Since the action of each agent is independent from the other elementary agents, it is possible to inactivate some of them (**Pause/Work** buttons). The values of $\Delta_{pos}$ and $\Delta_{or}$, used to normalize the agent contributions, can also be modified in real-time (**Position** and **Orientation** buttons) to adapt the contribution to the scale of the environment or to the task to perform.

Our experience shows that the contribution of the human operator is important in the optimization process. Indeed, if the automatic agent process fails (which can be the case when the cone used in the *visibility* agent is in collision with the environment), the human operator can:

- give to the MAS intermediate targets that will lead to a valid solution;
- move the manikin to a place where the MAS process could find a solution.

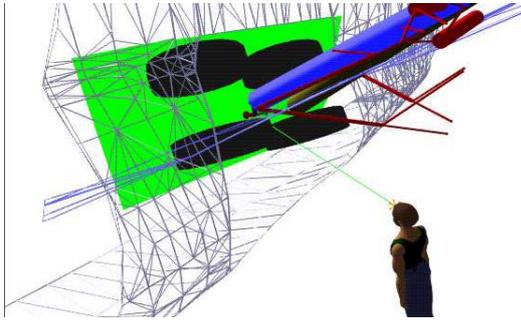

**Figure 10.** Digital model of a trap of an aircraft.

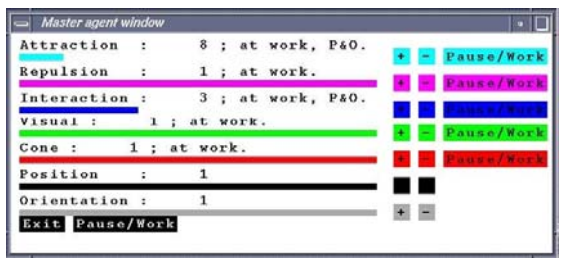

**Figure 11.** Master agent window.

On the other hand, algorithm tools allow the human operator to evolve more quickly and more easily with the DMU. The elementary agents guarantee a good physical and visual comfort and enable to quantify and qualify it, which would be a hard task for the human operator, even with sophisticated virtual reality devices. For instance, we can evaluate the rotations of the head and see how they are dispersed from a neutral configuration, inducing little effort.

The advantage of the MAS is to enable the combination of independent elementary agents to solve complex tasks. Thus, the agents participating in the visibility task can be coupled with agents enabling accessibility and maintainability as proposed by Chedmail [7]. That will be the purpose of our further works, as well as some developments to take into account the optimum distance to see the target according to the task to perform.

## Acknowledgments

This work has been experimented in the framework of the European project ENHANCE, acronym for "ENHanced AeroNautical Concurrent Engineering".